\documentclass[review]{elsarticle}

\journal{Pattern Recognition}

\usepackage{graphicx}
\usepackage{comment}
\usepackage{amsmath,amssymb} 
\usepackage{color}
\usepackage{epsfig}
\usepackage{xcolor}
\usepackage{subfigure}
\usepackage{enumerate}
\usepackage{algorithm}  
\usepackage{algorithmic}
\usepackage{multirow}
\usepackage{array}
\usepackage{enumitem}
\usepackage{helvet}  
\usepackage{courier}  
\usepackage{url}
\usepackage{longtable}

\usepackage{xspace}

\newcommand{\RV}[1]{{{#1}}}

\DeclareRobustCommand\onedot{\futurelet\@let@token\@onedot}
\def\@onedot{\ifx\@let@token.\else.\null\fi\xspace}

\def\eg{\emph{e.g}\onedot} 
\def\ie{\emph{i.e}\onedot}

\makeatother

\begin{document}

\begin{frontmatter}

\title{Spatial Feature Mapping for 6DoF Object Pose Estimation}


\author[mymainaddress]{Jianhan Mei}
\ead{jianhan001@e.ntu.edu.sg}

\author[mymainaddress]{Xudong Jiang}
\ead{exdjiang@ntu.edu.sg}

\author[mymainaddress]{Henghui Ding\corref{mycorrespondingauthor}}
\ead{ding0093@ntu.edu.sg}
\cortext[mycorrespondingauthor]{Corresponding author}

\address[mymainaddress]{School of Electrical and Electronic Engineering, Nanyang Technological University, Singapore}

\begin{abstract}
  This work aims to estimate 6Dof (6D) object pose in background clutter. Considering the strong occlusion and background noise, we propose to utilize the spatial structure for better tackling this challenging task. Observing that the 3D mesh can be naturally abstracted by a graph, we build the graph using 3D points as vertices and mesh connections as edges. We construct the corresponding mapping from 2D image features to 3D points for filling the graph and fusion of the 2D and 3D features. Afterward, a Graph Convolutional Network (GCN) is applied to help the feature exchange among objects' points in 3D space. To address the problem of rotation symmetry ambiguity for objects, a spherical convolution is utilized and the spherical features are combined with the convolutional features that are mapped to the graph. Predefined 3D keypoints are voted and the 6DoF pose is obtained via the fitting optimization. Two scenarios of inference, one with the depth information and the other without it are discussed. Tested on the datasets of YCB-Video and LINEMOD, the experiments demonstrate the effectiveness of our proposed method.
\end{abstract}


\end{frontmatter}


\section{Introduction}
\label{sec:introduction}

In this work, we address the challenging task of estimating object 6D pose from the image. It aims at recovering the 6D pose of each object instance in an image. More specifically, we focus on the rigid object 6D pose that contains 3 rotation parameters $(yaw, pitch, roll)$ and 3 translation parameters $(x, y, d)$ along 3 axes in the 3D coordinate system.

Object 6D pose estimation is a fundamental computer vision task in many applications, \eg robot manipulation and autonomous driving. Like many other computer vision tasks, pose estimation faces typical challenges, such as occlusion among object instances, background clutter, dynamic changes in the environment. In recent years, the success of deep neural networks~\cite{Recent2018Gu,liu2021few,sun2021m2iosr,lin2022flow,wang2021adaptive} in computer vision tasks has greatly promoted the development of pose estimation and has achieved many impressive achievements~\cite{SegDrivenYinlin, PoseCNN, MahdiBB8, DenseFusionChen,wang2021discovering}. Regrettably, most methods either directly regress 6D parameters based on image features, or estimate poses from the corresponding keypoints for every single object. Therefore, the problem of rotation symmetry and the 3D positional relationship of objects' points still cannot be well addressed.

\begin{figure}[ht]
    \centering
    \scalebox{0.7}{\includegraphics{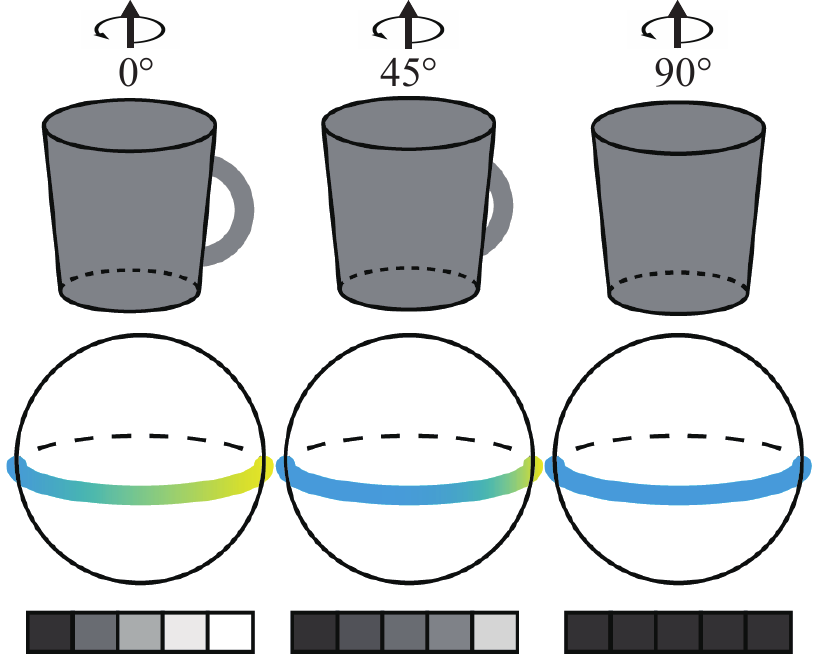}}
    \caption{The main idea of our spherical correlation method is to learn a latent spherical feature representation for each object that is consistent with the object appearances under different views so that the rotation on the object leads to the same rotation on its spherical feature. Here the views of an object (in the first row) are indexed by the rotations (in the second row) on $SO(3)$.}
    \normalsize
    \label{fig:motivation}
\end{figure}

\RV{The challenge of the rotation symmetry comes from similar object image appearances under different object rotations, \eg a symmetric object labeled as different poses may have very similar or even the same image appearance. In the learning stage, such data and labels bring out the problem of one versus multiple mappings, which makes the direct regression of the pose parameter an ill-posed problem. This results in poses with similar appearances ambiguous. As illustrated in Fig.~\ref{fig:motivation}, the handle of the cup indicates the rotation of the cup. However, during changing the views, with the disappearance of the handle, the cup with different rotations will have the same appearance.  This symmetry problem can be considered as a subset of the occlusion cases called self-occlusion, where the object occludes part of itself. Such pose ambiguity of symmetric objects may frequently occur in estimation, which is ill-posed problem as one versus multiple mappings is unsolvable by optimization.}
Efforts have been made on this symmetry problem. In~\cite{Implicit3DOriMartin}, the causes of pose ambiguities are grouped into object self-symmetry and occlusion-induced symmetry. Motivated by the recent advance on spherical convolutional neural networks~\cite{SphericalCNNTaco, LearningEquCarlos}, we propose to learn a latent spherical feature representation to object as an auxiliary for the pose estimation. The rotation parameter is on the Special Orthogonal ($SO(3)$) group which is consistent with the spherical correlation result so that the estimation of the rotation parameter can be proposed to be done by analyzing the spherical correlation defined in~\cite{SphericalCNNTaco} between the object appearance feature and the learned sphere feature representation. The object appearance feature is mapped to a hemisphere representing its one-side view. In Fig.~\ref{fig:motivation}, the latent spherical representation is required to be with the same rotation property of the object. By finding the maximum correlation between the object appearance spherical feature and the latent spherical representation, we do not confuse the network learning to predict different rotation parameters from similar appearances. During the inference, we pick the rotation parameter that has the largest spherical correction, which does not have to be the ground-truth rotation parameters as long as the corresponding projection of the 3D model well aligns with the ground-truth projection.

Furthermore, in the camera coordinate system, traditional methods for estimating the object pose typically utilize hand-crafted features to find the correspondence between the camera image and a predefined 3D model. The pose parameters are then calculated according to the correspondence. Due to the common shortcomings of hand-crafted features (\ie. SIFT~\cite{SIFTLowe}), such methods are very sensitive to the texture transition and lighting changes. Accordingly, the popularity of data-driven methods has brought new opportunities and challenges to pose estimation based on learning. More specifically, Deep Neural Networks (DNNs) have achieved remarkable success in many fields~\cite{zhang2021prototypical,zhang2021meta,ding2019boundary,ding2020semantic,liu2019feature,cai2021unified,li2021else}. For example,~\cite{PoseCNN, 3DRCNNAbhijit, mei2020object,3DPoseRegressionSiddharth} propose that the pose parameters are directly regressed from the convolutional features of each object. In this case, such methods are going to achieve significant improvements if there are sufficient training samples. However, one of the main challenges of pose estimation is background clutter, where objects may be surrounded by complex backgrounds or occluded by other objects. Noise will be mixed into the object's convolutional features, which makes it difficult for the regressor to fit the data according to the feature representation. To overcome this problem, a method of obtaining translation from convolutional feature maps by voting is first proposed in PoseCNN~\cite{PoseCNN}.

Recently, inspired by the traditional methods of inferring pose parameters using the correspondence, ~\cite{SegDrivenYinlin, PVNetSida, Car3DBBox} define and calculate the keypoints of the objects, and then find the pose based on the corresponding keypoints through the "Perspective-n-Point" (PnP) algorithm. Usually, the keypoints of these methods are obtained through a coordinate map according to the surface of the object in the image, and the keypoints are determined by voting of all pixels from the coordinate map. The advantage of this is not only that the keypoints prediction is pixel-by-pixel, which can handle occlusion and background noise in features, but also that voting effectively suppresses outliers. Following the above-mentioned pose estimation idea based on keypoints, \cite{SegDrivenYinlin, PVNetSida, Car3DBBox, CDPN2019Zhigang} predict the coordinate map directly from the 2D convolutional features where 3D information is insufficiently used. Moreover, \cite{DenseFusionChen, PVN3D2020Yisheng} merge the 2D features with 3D points~\cite{li2022primitive3d}, and the PointNet~\cite{PointNet2017Charles} is applied for further calculation. However, in most of the keypoint-based methods, the 3D relation information is not sufficiently considered. Although PointNet~\cite{PointNet2017Charles} is good at dealing with disordered points, the Multilayer Perceptron (MLP) may destroy the essential structure of the data.~\RV{6D object pose estimation is closely related to 3D geometry measurement. Among different coordinate spaces, the 6D pose parameter is one of the essential descriptions for objects in 3D space. There are a few ways to build the essential representation for a 3D object, \eg, point cloud, voxel, mesh and 3D surface. While voxel, mesh and the 3D surface can easily store the 3D relationship between elements, the relation information is hard to be processed by neural networks. PointNet~\cite{PointNet2017Charles} and O-CNN~\cite{OctreeCNNWangLGST17} provide solutions for neural networks to deal with point cloud and voxel data respectively while Graph Convolutional Networks (GCNs)~\cite{3DHand2019Liuhao} make it possible for graph data inference by a neural network. While point cloud is leaking the 3D connection and voxel is redundant, mesh is considered a more elegant representation for a neural network with spatial relation information. Moreover, mesh is one kind of special graph that uses its nodes as vertices and connections as edges. Hence, instead of using point cloud or voxel, we connect discrete points into mesh to build a graph model of a 3D object. Next, GCN is used to help the feature exchange and merge. Finally, the predefined key points are regressed and voted from the vertices of the graph, and the pose parameters are calculated through optimization. Using the mesh representation for data inference provides the system with a nature graph scheme. Compared with MLP in PointNet~\cite{PointNet2017Charles}, GCN considers more on the connection relationship and data spatial topology. Using GCN combined with the mesh representation sheds light on building a union 3D data representation for neural network inference.}

Our contributions are summarized as follows:
\begin{itemize}
\setlength\itemsep{0em}
\item We explore to solve the rotation ambiguous and occlusion problems in 6D pose estimation by using the spherical correlation and graph convolution. A robust 6D object pose estimation system is proposed.

\item We propose to map the 2D convolutional feature to both a sphere and the 3D mesh representation. And the corresponding network components are integrated which forms the end-to-end deep neural network training and inference scheme. The proposed method digs more essential information from the data representation and processing. To our knowledge, this is new and meaningful for 6D object pose estimation.

\item Target on the rotation ambiguous and occlusion challenges, we propose to solve from the parameter space and data structure. The state-of-the-art performance demonstrates the efficiency of the proposed system.
\end{itemize}

\section{Related Work}
\label{sec:related_work}

\subsection{6D Object Pose Estimation}
Early methods focus on recovering poses by matching keypoints features between 3D models and images~\cite{MOPEDAlvaro, SIFTLowe, LocalAffineFred, Learning2015Ren, Interest2013Miao}. Correspondences are found through the matching and the parameters are recovered by further optimization. However, these methods are limited by hand-crafted features and suffer from the problems of the keypoints extraction and description, which are not robust to \eg illumination changing, non-significant texture and affine transform. Thus, with additional depth information, \cite{ModelTextureStefan, Learning6D3DObjCoorEric, li2021mine,LearningHierarchicalSparseLiefeng, RGBDPretrainedMax, DLLocalPatchWadim} significantly improve the accuracy performance.

Recently, Deep Neural Networks (DNNs) have achieved remarkable success in many computer vision fields~\cite{ding2018context,ding2021vision,ding2019semantic,shuai2018toward,tong2021directed,wang2019bi,wanglearning,mei2019deepdeblur,liu2022instance,liu2021towards,chiou2021recovering,ding2020phraseclick,wang2019dermoscopic}. The state-of-the-art 6D object pose estimation methods from monocular images are mostly based on DNNs. For example, convolutional features bring significant performance improvements~\cite{3DRCNNAbhijit, FastSSDPoseEstPatrick, Car3DBBox}. More specifically, PoseCNN~\cite{PoseCNN} proposes a representation that infers the actual 3D coordinates of the image coordinates then determines the object position center through Hough voting and designs ShapeMatch-Loss to solve the problem of rotationally symmetric object pose estimation. DeepIM~\cite{DeepIMYiLi} proposes a method to predict the relative translation and relative rotation of objects in two images so that the network can learn how to fine-tune the posture. In addition, it also proposes an untangled relative pose representation for accurate pose prediction. Unlike the pose estimation methods based on object detection, this representation can handle previously unseen objects. Besides, considering the 3D spatial properties of the object pose, more additional supervision information is used to enhance the performance of deep neural networks. \cite{SegDrivenYinlin, PVNetSida} learn the keypoints of 2D to 3D matching through the instance segmentation framework to enhance the description of the object pose, which brings the system performance to a new level. However, these methods all focus on utilizing the 2D features for better regression, where the 3D information may be insufficiently used. Thereupon, \cite{DenseFusionChen, PVN3D2020Yisheng} propose to do the fusion of 2D and 3D features. 2D points are sampled from the convolutional feature map. With the depth map, 3D coordinates are recovered from the sampled 2D points. The feature fusion is typically the concatenation of the 2D convolutional features and the 3D coordinates. Network inference in 3D space achieves better interpretation for the pose parameters and reaches more accuracy.

Note that \cite{Implicit3DOriMartin, Fabian2019ExplainingICCV} specifically aim to address the symmetry ambiguous problem. Both of them try to learn a feature representation from multiple object views, and the final rotation is obtained by searching the feature space. Different from their methods, we propose to learn the object representation on a sphere which is an essential representation for object rotation and we do not require multi-view images.

\subsection{Symmetric and Spherical Correlation}
\RV{It is not difficult to understand that symmetric objects cause ambiguity in mapping appearance to pose. As aforementioned, the symmetric ambiguity makes the learning-based method into a one versus multiple mapping task, which is an ill-posed problem without a unique solution. By learning from such an ill-posed problem, the results fall into the unpredictable. Using such estimation for higher-level tasks may cause the system out of control. Recently, efforts have been made to tackle this problem. One intuitive solution is to do symmetry detection or analysis for each specific object. Efforts have been made on both extrinsic symmetry~\cite{PartialNiloy, DetsymmDilip, SymmetryPriorPablo} and intrinsic symmetry~\cite{FullPartialDan, FieldsDaniele, BlendedVladimir, ApproximateChuan, NautilusMichal} detection in 3D space~\cite{SymmetryDetRajendra}. However, such methods mainly focus on the symmetry of 3D models and often ignore the object image appearances under specific views, which are important for the pose parameter regression task.}

In~\cite{3DRCNNAbhijit}, a render and compare loss is introduced to compare the rotated and rendered 3D model with its segmentation and depth masks. Penalized by the Intersection over Union (IoU) between the rendered and the label masks, the ambiguous object poses are represented as hidden parameters in the network that can have the same appearance. However, solving the hidden parameters is still ill-posedd. Moreover, such render and compare loss only focuses on the rendering shape but ignores the object's texture. In~\cite{Implicit3DOriMartin}, an encoder is learned that encodes the object poses in a feature codebook and the testing pose is obtained by retrieving from the codebook. However, learning the encoder needs training samples from all object rotation views. The rotation of an object on the $SO(3)$ group is difficult to be uniformly sampled. In addition, the code distance for describing different rotation views is hard to define.

Spherical signals are a set of particular data with special properties on the sphere. It is known that the flat 2D convolution is a translation equivariance operation. Similarly, it is proved that the convolution on the sphere and $SO(3)$ is equivariance on the rotation group $R{\in}SO(3)$~\cite{SphericalCNNTaco}. Particularly, in~\cite{SphericalCNNTaco, LearningEquCarlos}, the spherical convolution on a sphere and $SO(3)$ is defined and discussed. With that, we propose to learn a rotation equivariance spherical feature representation for 6D object pose estimation.

\subsection{Graph Convolutional Network}
GCN has demonstrated its strength in many fields. In 3D representation, the graph has been widely used. \cite{SGCN2019Long, Spatial2018Sijie} explore the application of GCN on the human body and hand joints, while \cite{3DHand2019Liuhao} reconstruct the complete 3D mesh for the human hand using GCN. Considering that mesh and surface are popular 3D representations, they are all consisted of points and connections, which can be naturally abstracted as graphs. In~\cite{wang2018pixel2mesh, wang2019pixel2meshpp}, the graph is built based on the 3D points and mesh connections, and GCN is used for the graph inference. The 3D shape generation is changed to the task of deformation from an initialized shape to the target shape. However, the inference of the graph is also a process of regression that predicts the target point coordinate on each node. From this view, \cite{HOPENet2020Bardia} GCN is used to jointly learn the hand-object pose, where the object pose is described by its keypoints.

\section{Methodology}
\label{sec:method}
\begin{figure*}[h]
    \centering
    \scalebox{0.65}{\includegraphics{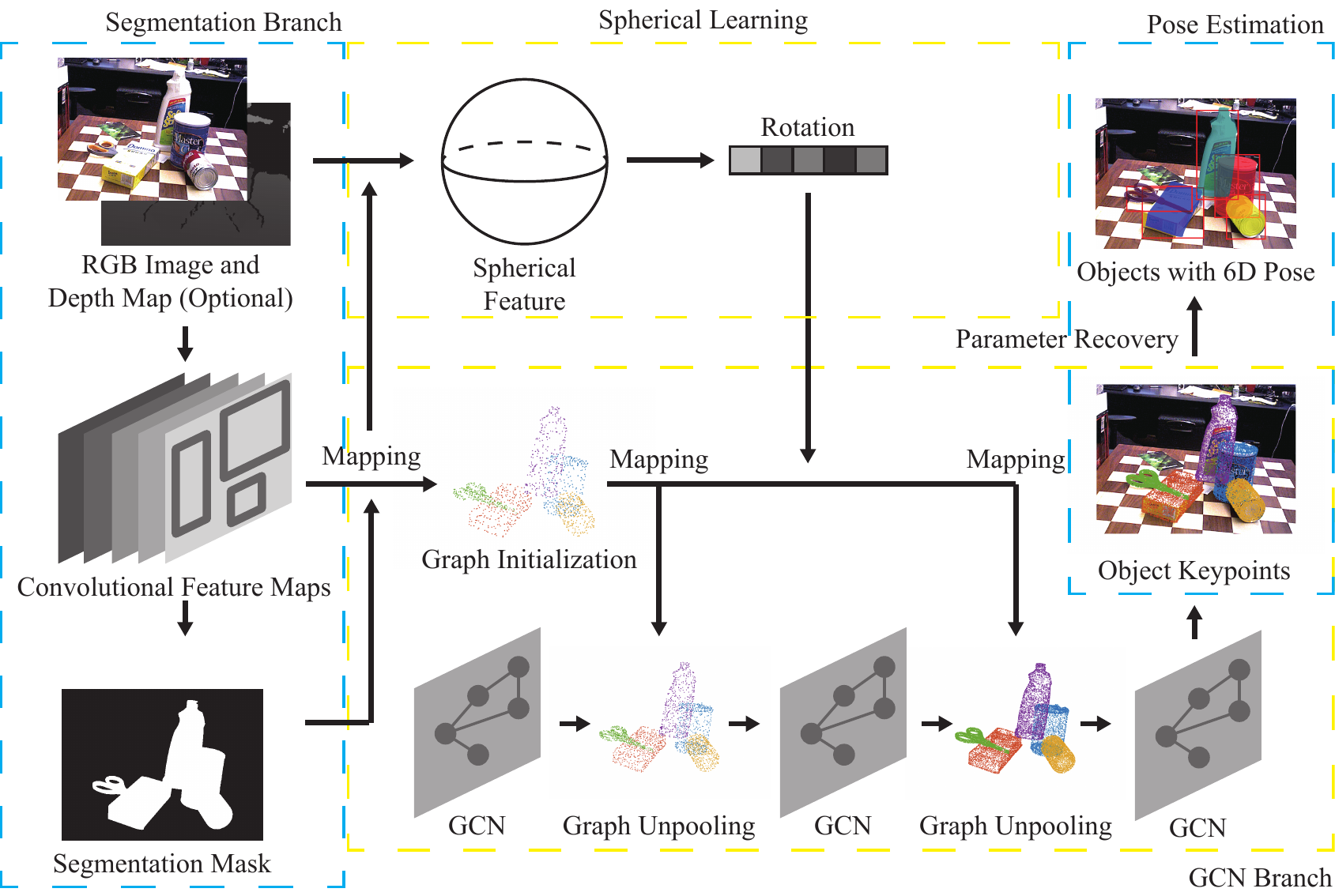}}
    \vspace{-3mm}
    \caption{The overall pipeline of the proposed method. Our system is divided into four parts, which are a segmentation branch for feature extraction and sampling, a spherical learning branch for coarse rotation prediction, a GCN branch for capturing the 3D object structure and inferring the graph with 2D features, and a pose estimation utilizing the voted keypoints for parameter regression. The system takes an RGB image and a depth map as input, where the depth map is not compulsory. Our core data structure is a graph abstracted according to the 3D mesh connection. The graph is filled by 2D to 3D correspondence and filled with 2D convolutional features and spherical features. During the graph inference, unpooling will be applied so that the model can be better learned and there are enough output points to support the keypoint voting. In the GCN branch, two graph unpooling layers connect 3 groups of GCN layers. The posture parameters are calculated by optimizing the fitting function.}
    \normalsize
    \label{fig:overall_struct}
\end{figure*}

As shown in Fig.~\ref{fig:overall_struct}, a monocular RGB image with its depth map is taken as the input of our overall pipeline, where the depth map is optional. The object masks and their Convolutional Neural Network (CNN) features are obtained in a semantic segmentation branch~\cite{ding2021interaction,ding2022deep}. A coarse rotation is learned by spherical learning. Inspired by~\cite{wang2018pixel2mesh}, a graph is abstracted according to the object mesh formed by all the seen object surfaces. Afterward, the corresponding convolutional features combining the spherical feature are mapped to 3D coordinates and further refined through a graph matching algorithm. The Graph Convolutional Network (GCN) is used to process the graph model and a set of predefined 3D object keypoints are predicted. The graph unpooling layer is introduced so that the graph scale can be well controlled and the graph output can be dense enough to support the keypoint voting. The final pose parameters are obtained by optimization according to the keypoints.

\subsection{Spherical Representation Learning}
For the self-contained purpose, here we introduce some important concepts for spherical signals and operations on a sphere as well as on the $SO(3)$ group introduced in \cite{SphericalCNNTaco, LearningEquCarlos}.

Without losing generality, we only introduce the continuous version of the spherical operations, which can be easily converted to the discrete case. Mainly following the notations in~\cite{SphericalCNNTaco}, similar to a 2D image, a sphere signal is defined as a continuous function on the spherical coordinates $f:S^2{\rightarrow}{\mathbb{R}}^K$, where $S^2$ indicates the spherical coordinates and $K$ is the dimensionality of the continuous function. Rotating a spherical signal with a rotation operator $O_R$ indexed by $R{\in}SO(3)$ has the property:
\begin{equation}
[O_Rf](x)=f(R^{-1}x),
\label{equ:rotation_spherical_signals}
\end{equation}
where the $R^{-1}$ is the inverse of $R$. \eqref{equ:rotation_spherical_signals} essentially states rotating a sphere signal by $R$ is equivalent to rotating the sphere coordinate system by $R^{-1}$.

Considering that the normal 2D convolution can be defined as the function of the inner product over the 2D plane, the inner product of two spherical signals is defined as:
\begin{equation}
[\phi*f]=\int_{S^2}\sum_{k=1}^{K}{\phi_k(x)}{f_k(x)}dx,
\label{equ:spherical_inner_product}
\end{equation}
where $\phi$ is another spherical signal with the same dimensionality of $f$. Different from the inner product on the 2D plane, the inner product on sphere calculates the integration along sphere $S^2$ which is indexed by the spherical coordinates $\alpha$ and $\beta$. Its integration measure $dx$ is a standard rotation invariant one, $d{\alpha}sin({\beta}/4{\pi})$, which ensures $\int_{S^2}f(Rx)dx=\int_{S^2}f(x)dx$ for any rotation $R{\in}SO(3)$. Further, the spherical correlation is defined as~\cite{SphericalCNNTaco}:
\begin{equation}
[\phi*f](R)=\int_{S^2}\sum_{k=1}^{K}{\phi_k(R^{-1}x)}{f_k(x)}dx,
\label{equ:spherical_correlation}
\end{equation}
where $R{\in}SO(3)$ is the index of the spherical correlation.

The spherical correlation can also be defined on $SO(3)$~\cite{SphericalCNNTaco}:
\begin{equation}
[\phi*f](R)=\int_{SO(3)}\sum_{k=1}^{K}{\phi_k(R^{-1}Q)}{f_k(Q)}dQ,
\label{equ:SO3_correlatoin}
\end{equation}
where $\phi$ and $f$ now denote functions defined on $SO(3)$, and $dQ$ denotes the invariant integration measure on $SO(3)$. Under ZYZ-Euler angles, $(\alpha, \beta, \gamma)$, $dQ$ becomes $d{\alpha}sin({\beta})d{\beta}d{\gamma}/(8{\pi}^2)$~\cite{SphericalCNNTaco}. Both spherical correlations in \eqref{equ:spherical_correlation} and \eqref{equ:SO3_correlatoin} are rotation equivariance, \ie.
\begin{equation}
[\phi*O_Qf](R)=[\phi*f](Q^{-1}R)=[O_Q[\phi*f]](R).
\label{equ:equivariance}
\end{equation}

This property states that the result from a sphere correlation preserves the original signal property on $SO(3)$.

This step aims to estimate coarse rotation parameters on $SO(3)$. Our basic idea is to map the segmentation feature of an object onto a hemisphere. If there is a spherical representation of the object, we can use the spherical correlation to generate the correlation output on $SO(3)$. Since we have the ground-truth 3D model for each object, a straightforward way is to follow~\cite{SphericalCNNTaco} to use ray casting to map a 3D model to a sphere. Specifically, towards an origin, a ray shots from each point on the sphere surface and ends where it hits the model surface. The ray length and surface angle of each point are recorded and form the spherical signal. Using the ray casting sphere $f$ as the reference spherical representation of the object, the object rotation can be found by learning the object feature under each specific view that has the maximum correlation with $f$. For a specific symmetric 3D model, when the ray origin coincides with its symmetry center, the ray casting spherical signal has the same symmetry attribute. However, the ambiguous poses come from both the object self symmetry and the occlusion induced symmetry. It is hard to find the symmetry origin for occlusion induced symmetry by only using the 3D model.

\begin{figure}[h]
    \centering
    \scalebox{0.9}{\includegraphics{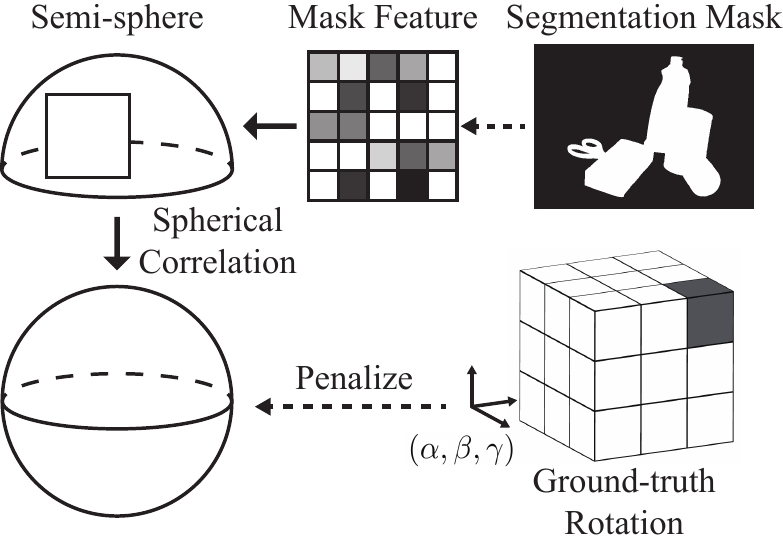}}
    \vspace{-3mm}
    \caption{The segmentation feature is mainly learned from the ground-truth mask, and the spherical representation is mainly learned according to the ground-truth rotation.}
    \vspace{-1mm}
    \normalsize
    \label{fig:spherical_learning}
\end{figure}

Thus, instead of using the ray casting sphere $f$, we propose to learn an auxiliary spherical representation $f$ that has the same symmetry property as the object, to handle the pose ambiguity. Specifically, the spherical representation $f$ is represented by a branch of learned object-specific parameters of a multi-layer SphereCNN~\cite{SphericalCNNTaco}, where its input is the segmentation feature of an object projected to a hemisphere and its output is the spherical correlation on the angle grid sampled on the rotation angles on $SO(3)$, as illustrated in Fig.~\ref{fig:overall_struct}.

In particular, we use the "Driscoll-Healy" grid following~\cite{DriscollComputing, ChirikjianEngineering} to project the segmentation feature of an object to a hemisphere denoted as $\phi$. The spherical representation $f$ or a branch of the SphereCNN consists of 5-layer parameters, where the first layer performs the spherical correlations on $S^2$ defined in Eq.~\ref{equ:spherical_correlation} and the rest four layers perform the spherical correlations on $SO(3)$ defined in Eq.~\ref{equ:SO3_correlatoin}. The rotation equivariance property specified in Eq.~\ref{equ:equivariance} ensures the spherical correlation result to have the same rotation property as the object.

To learn the unknown spherical representation $f$, the segmentation features $\phi$ consistent with the object view appearance are required. As shown in Fig.~\ref{fig:spherical_learning}, the hemisphere segmentation feature $\phi$ is mainly learned from the ground-truth segmentation mask and the unknown spherical representation $f$ is mainly learned from the target rotations.
We formulate learning the auxiliary spherical representation as minimizing the following target:
\begin{equation}
f^* = \mathop{min}\limits_f{\left\| {R_g - \mathop{argmax}\limits_{R_p}([\phi*f](R_p))} \right\|},
\label{equ:spherical_learning}
\end{equation}
where $R_p$ and $R_g$ are the predicted and ground-truth rotations, respectively. Eq.~\ref{equ:spherical_learning} aims at learning the spherical representation $f^*$ by minimizing the difference between $R_p$ and $R_g$, where $R_p$ is obtained by finding the rotation that maximizes the spherical correlation.

During training, the spherical representation is updated at each iteration mainly supervised by the ground-truth rotation $R_g$ with a cross-entropy loss:
\begin{equation}
{L_{s}} = -log([\phi*f](R_g)).
\label{equ:correlation_loss}
\end{equation}

Note that the spherical correlation result needs to be normalized first before computing the loss. Finally, we take a coarse rotation estimation by maximizing the spherical correlation result. In detail, we do max-pooling on the features from the spherical correlation, which is different from~\cite{SphericalCNNTaco} that uses the $SO(3)$ integration to obtain the full rotation invariance. The max-pooling on $SO(3)$ space makes the inference sparse. The network goes to non-convergence when the rotation supervision is unavailable.

\subsection{Feature Sampling}
Observing the depth map containing noise, considering the scenario where the depth map is unavailable, the mapping from the image to the 3D graph is based on 2D sampling. Inside the object's segmentation mask, Poisson disc sampling~\cite{bridson2007fast} is applied and a fixed number of coordinates are collected.

During training, the segmentation branch is supervised by semantic labels. The feature maps before the last output are used for feature extraction. The coordinates collected through Poisson disk sampling~\cite{bridson2007fast} are rescaled into feature maps from different layers according to their times of pooling and unpooling. Since the collected coordinates are continuous, linear interpolation is applied along the feature channel. After the sampling of the 2D convolutional feature maps, the spherical features are sampled according to their 2D correspondences. The 2D convolutional features are combined with the spherical features which will be fed to the next stage.

\subsection{Graph Building and Filling}
Assume that a 3D mesh is a collection of vertices and edges that defines a 3D structure, it can be represented as a graph model $G = (V, E)$. $V$ is the set of nodes, and $E$ is the set of edges. The graph convolutional layer can be written as a nonlinear function~$H^{l+1} = f(H^l)$, where $H^0 = X$ is the input of the first layer, $X \in R^{N \times D}$, $N$ is the number of nodes in the graph, $D$ is the dimension of the feature vector of each node. According to this definition, our method relies on~\cite{bronstein2017geometric} to handle 3D geometry. On an irregular graph, a graph convolutional layer is defined as:
\begin{equation}
H^{l+1}_{u} = \sigma (w_0 H^l_u + \sum_{v \in A(u)} w_1 H^l_v),
\label{equ:gcn_layer}
\end{equation}
where $u$ and $v$ are the vertices of the graph model, $\sigma (\cdot)$ denotes a nonlinear activation function, $A(u)$ indicates the neighboring vertices of $u$, $w_0$ and $w_1$ are learnable parameter matrices, and $w_1$ is shared for all edges.

In~\cite{DenseFusionChen, PVN3D2020Yisheng}, the 3D points are corresponding to depth map pixels which are aligned with the image. In our system, GCN is more flexible where graph pooling and unpooling can be applied. Because the graph is required to cover all objects in the image, the number of vertices of the graph may not be the same as the 2D sampling features. We thus consider two cases of constructing the graph with and without depth maps. Thereupon, a feature merging method based on graph matching is proposed.

\begin{figure}[h]
    \vspace{-1mm}
    \centering
    \scalebox{0.9}{\includegraphics{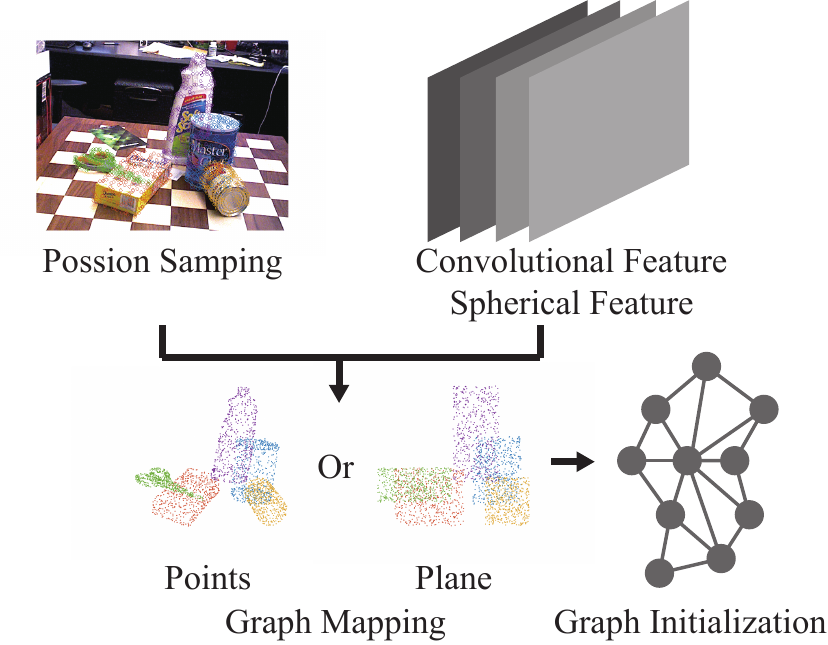}}
    \vspace{-3mm}
    \caption{Our graph can be initialized by either object 3D points or a plane with average depth if the depth map is unavailable.}
    \vspace{-1mm}
    \normalsize
    \label{fig:FeatureMapping}
\end{figure}

For each instance point sampling and mapping, as shown in Fig.~\ref{fig:FeatureMapping}, we recover the 3D point of the object through each pixel and the known intrinsic matrix if the depth map is available. We define a square region with a set of uniformly distributed nodes as a regular grid. By connecting the nodes, the graph is initialized and is re-scaled as the inscribed polygon of the sampled 2D features. The static graph is used for the inference in which the graph structure is fixed but the features are filled to it. For the mapping from the 2D features to the graph, each vertex is first connected to the top $N$ closest 2D points on the image. Then, consisting of a bipartite graph by the 3D points connected to the 2D points, we apply graph matching to make the features better distributed on the sphere. Finally, each 3D point retains only one assigned connection, and the corresponding 2D features are filled to the vertices.

If there is no depth map, it is impossible to directly recover the 3D points of the object. Similarly, we will construct the graph by sampling uniformly on the segmentation regions. Afterward, fill the vertices by finding the top N closest points of the vertices. To initialize 3D positions of the graph vertices, an average depth of the training dataset is used and the graph is initialized as a plane. Finally, the graph is also refined through graph matching.

For building the inference graph, a bipartite graph matching mechanism is used. Considering that the graph network is much harder going to convergence, the static graph is built. Meanwhile, for better training the network, the 2D features are sampled by Poisson disk sampling that introduces randomness to the features. During building the graph, the extracted feature points are assigned to their top $N$ closest 3D mesh. To reduce the inconsistency between the static graph and randomly sampling, the bipartite graph matching mechanism is utilized for the feature merging. After the bipartite graph matching, the repeat assignments are cleared and the unique mapping is preserved.

Finally, according to the assignment, the 2D features from the convolutional network are extracted and filled into the graph vertices. Depending on the availability of the depth information, the vertex coordinates are initialized by either the plane with average depth or the points with depth. The final feature in each vertex is the concatenation of the convolutional feature and the vertex coordinate. After connecting the edges by the triangle mesh, the graph is built.

\subsection{Graph Unpooling}
A 3D mesh with rich details requires a large number of points to represent. As its abstracted representation, the graph complexity grows exponentially with the increasing number of mesh points. For better inference and reducing the data redundancy, graph unpooling is introduced into our network.

\begin{figure}[ht]
    \centering
    \scalebox{0.8}{\includegraphics{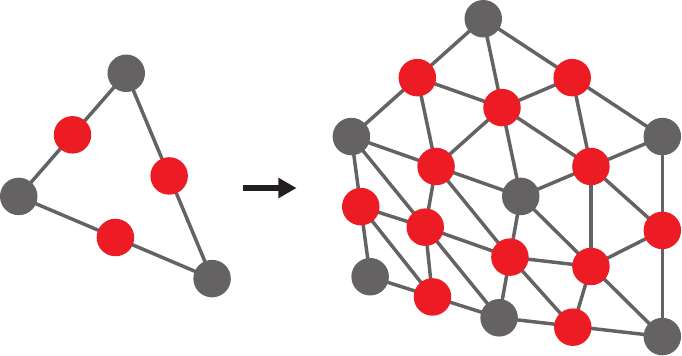}}
    \caption{New vertices are first inserted in the middle of every edge. And then the new vertices (nodes) are connected by the Delaunay tesselation method.}
    \label{fig:Unpooling}
\end{figure}

The graph is initialized with fewer vertices. After several unpooling layers, a sufficient number of vertices can be obtained. As indicated in Fig.~\ref{fig:Unpooling}, new vertices are inserted in the middle of every edge. Then new features are interpolated by two headers on each edge. Different from~\cite{wang2018pixel2mesh} where new connections are inside each triangle, after the interpolation, we rebuild the mesh by the Delaunay tesselation method for all the vertices. During the network inference, multiple stages of unpooling will be performed. After each graph unpooling, features from the convolutional layers will be mapped and concatenated to the new graph, which exploits the information from low-level layers. Finally, the high-level graphs mix the feature from 2D convolution maps and the GCN inference.

In this way, the graph scale can be well controlled, while the graph network will easily go to convergence. Moreover, in the proposed system, the 3D shape is utilized for supervision. The dense output points are able to better fit the shape label, which helps the optimization of the network. Following the interpolation rule of unpooling, the number of points grows quickly. In our system, two graph unpooling layers are enough for obtaining a sufficient number of the final output points that provide further keypoint voting.

\subsection{Losses}
To better capture the spatial information of the 3D objects, in addition to predicting keypoints, the graph deforms the surface points of the object at the same time. During training, the graph is mainly supervised by the ground-truth object model points and the keypoints.

Graphic deformation helps to find the surface of the object more precisely. There is no point in calculating the absolute distance between two point sets. To measure the similarity of two distributions, Chamfer distance is more appropriate than the absolute distance~\cite{LocatingJavier, AveHausdorffOliver, Metrics3DMedAbdel}. Letting $A$ and $B$ denote as the two point sets, the Chamfer distance is defined as follows:
\begin{equation}
d_{CD}(A,B) = \frac{1}{\left| A \right|}\sum_{a\in{A}}\mathop{min}\limits_{b\in{B}}d(a,b)+\frac{1}{\left| B \right|}\sum_{b\in{B}}\mathop{min}\limits_{a\in{A}}d(a,b),
\label{equ:ave_Chamfer}
\end{equation}
where $d(a,b)$ denotes the distance between the two points $(a,b)$, which is $L_2$ distance in our work, and $\left| A \right|$ and $\left| B \right|$ denote the numbers of points in sets $A$ and $B$, respectively. Thus, the Chamfer loss is computed using the distance.
\begin{equation}
L_{CD} = d_{CD}(P_{init},P_{init}+t_{offset})
\label{equ:Chamfer_loss}
\end{equation}
where $P_{init}$ are the points initialized in the graph vertices, and $t_{offset}$ is the predicted deformation offset.

For keypoint learning, each vertex on the graph predicts the offset from the vertex position to each keypoint. We mainly refer to the $L1$ loss in~\cite{PVN3D2020Yisheng} and configure it as follows:
\begin{equation}
L_K = \frac{1}{\left| P_c \right|}\sum_{p\in{P_c}}\sum_{k\in{K}}\left\lVert off(p,k) - off_{gt}(p,k) \right\lVert_{L1},
\label{equ:L1_loss}
\end{equation}
where $P_c$ is a set containing all the points belonging to category $c$. $K$ is the keypoint set. $off()$ and $off_{gt}(p,k)$ denotes the function computing the offset between the two points.

\section{Experiment}
\label{sec:experiment}

\subsection{Implementation Details}
Our system extracts features from the segmentation branch. The segmentation network is of the encoder-decoder structure. It outputs a semantic segmentation map with $N+1$ channels for $N$ object categories and $1$ negative channel. We directly use the trained segmentation network in~\cite{DenseFusionChen, Knowledge2021Wang} but fine-tune the convolutional feature for the graph inference.

The spherical learning branch is built by the filters in the spherical correlation that contains one layer of $S^2$ correlation and four layers of $SO(3)$ correlation. For the spherical learning, we use the angle grid size of 20 by default, although later we also test different grid sizes of the spherical correlation.

\begin{table*}[htbp]
\renewcommand\arraystretch{0.8} 
\footnotesize
\begin{center}
\begin{tabular}{l|c|c}
\hline
Layer Name  &Output Dim.  &Graph Nodes\\
\hline
\hline
GCN 1  &512+3  &400  \\
GCN 2  &192  &400  \\
GCN 3  &192  &400  \\
Graph unpooling 1   &192  &3652  \\
GCN 4  &192  &3652  \\
GCN 5  &192  &3652  \\
Graph unpooling 2   &192  &16147  \\
GCN 6  &3  &16147  \\
\hline
\end{tabular}
\end{center}
\vspace{-5mm}
\caption{GCN structure}
\label{Tab:GCNStructure}
\end{table*}

The graph convolutional network (GCN) is implemented fitting from~\cite{GCN2017Thomas}. The structure and parameter are shown in Table~\ref{Tab:GCNStructure}. The input graph has the feature dimension of 515 filled in 400 vertices. Expect for the first and the last one, the GCN layers are all of 192 dimension output. The last layer output provides the 3 dimensions of coordinates. First, a regular grid of $20\times20=400$ points is initialized. Then we build the undirected graph using the 3D points as vertices and their mesh connections as edges, where the mesh is formed by the Delaunay tessellation method. As mentioned by \cite{GCN2017Thomas}, the graph is represented by its adjacency matrix. Chebyshev polynomial coefficients are fitted based on the Laplace matrix. In the inference process, GCN mainly predicts Chebyshev polynomial coefficients, and the final output is obtained through an inverse transformation. After 2 stages of graph unpooling, the final output graph scale contains 16147 nodes.

Our GCN branch also predicts the 3D object shape though the graph is initialized by the depth map or the plane. The supervision is built by the transformed 3D object models. Only the surface facing the camera is extracted, and a fixed number of points are sampled for the Chamfer distance loss function.

Depending on if the depth information is available or not, the graph can be initialized by either the points from the depth map or the plane with average depth. The average depth is obtained by calculating the mean object depth in the dataset. Results from two different initialization are compared.

For keypoint selection, we follow the settings in~\cite{PVNetSida, PVN3D2020Yisheng}. Points on each object model are selected instead of predefined geometry points outside the model for more precise estimation. The output of the GCN represents the keypoints by the offsets from the transformed keypoints to the current model points. Then, the parameters are fitted according to the keypoints in a Hough Voting manner. The points are first transformed to their parameter space and inliers are used for the least-square fitting.

\subsection{Evaluation Metrics}
Most of the metrics for object 6D pose estimation can be divided into two categories. One focuses on the absolute precision of the pose parameters and the other focuses on measuring the intersection between the transformed or rendered object 3D models under the ground-truth and the predicted pose parameters, where the latter can consider the symmetry property of objects.
\cite{ModelTextureStefan, PoseCNN} introduce two metrics: Average Distance of Distinguishable (ADD) and its symmetry version, Average Distance of Distinguishable Symmetry (ADD-S). ADD computes the absolute distance between two transformed 3D model point sets:
\begin{equation}
ADD = \frac{1}{\left| V \right|}\sum_{v\in{V}}{\left| (R_gv + t_g) - (R_pv + t_p) \right|},
\label{equ:ADD}
\end{equation}
where $V$ denotes the 3D model point set and ${\left| V \right|}$ is the number of points in the set. $R_{g,p}$ and $t_{g,p}$ are the ground-truth and predicted rotation and translation. And ADD-S considers the symmetry case by replacing the absolute distance by an average minimum distance which is defined as:
\begin{equation}
ADD-S = \frac{1}{\left| V \right|}\sum_{v_1\in{V}}\mathop{min}\limits_{v_2\in{V}}{\left| (R_gv_1 + t_g) - (R_pv_2 + t_p) \right|}.
\label{equ:ADDS}
\end{equation}

Following~\cite{PoseCNN, DenseFusionChen, PVN3D2020Yisheng}, we calculate the Area Under Curve (AUC), the area under the threshold curve, for the two average distance, ADD and ADD-S. For YCB-Video, both the ADD and ADD-S AUC are compared. And only the ADD is computed for LINEMOD.

\begingroup
\footnotesize
\centering 
\renewcommand\arraystretch{0.88}
\begin{longtable}{@{\extracolsep{\fill}}l|c|c|c|c}

\hline
& \begin{tabular}{@{}c@{}}PoseCNN \\ (ICP)\end{tabular}  & \begin{tabular}{@{}c@{}}DF \\ (iterative)\end{tabular}  & \begin{tabular}{@{}c@{}}PVN3D \\ (ICP)\end{tabular} & \begin{tabular}{@{}c@{}}Ours \\ (depth)\end{tabular}  \\
\hline
\hline
002\_master\_chef\_can     & 95.8  & 96.4  & 95.2  & 93.2  \\  \hline
003\_cracker\_box          & 92.7  & 95.8  & 94.4  & 95.6  \\  \hline
004\_sugar\_box            & 98.2  & 97.6  & 97.9  & 98.3  \\  \hline
005\_tomato\_soup\_can     & 94.5  & 94.5  & 95.9  & 97.5  \\  \hline
006\_mustard\_bottle       & 98.6  & 97.3  & 98.3  & 97.9  \\  \hline
007\_tuna\_fish\_can       & 97.1  & 97.1  & 96.7  & 96.5  \\  \hline
008\_pudding\_box          & 97.9  & 96.0  & 98.2  & 98.3  \\  \hline
009\_gelatin\_box          & 98.8  & 98.0  & 98.8  & 97.8  \\  \hline
010\_potted\_meat\_can     & 92.7  & 90.7  & 93.8  & 93.3  \\  \hline
011\_banana                & 97.1  & 96.2  & 98.2  & 97.9  \\  \hline
019\_pitcher\_base         & 97.8  & 97.5  & 97.6  & 97.9  \\  \hline
021\_bleach\_cleanser      & 96.9  & 95.9  & 97.2  & 96.7  \\  \hline
024\_bowl*                 & 81.0  & 89.5  & 92.8  & \textbf{94.3}  \\  \hline
025\_mug                   & 94.9  & 96.7  & 97.7  & 97.6  \\  \hline
035\_power\_drill          & 98.2  & 96.0  & 97.1  & 95.9  \\  \hline
036\_wood\_block*          & 87.6  & 92.8  & 91.1  & \textbf{94.1}  \\  \hline
037\_scissors              & 91.7  & 92.0  & 95.0  & 95.5  \\  \hline
040\_large\_marker         & 97.2  & 97.6  & 98.1  & 98.1  \\  \hline
051\_large\_clamp*         & 75.2  & 72.5  & \textbf{95.6}  & \textbf{95.6}  \\  \hline
052\_extra\_large\_clamp*  & 64.4  & 69.9  & 90.5  & \textbf{91.1}  \\  \hline
061\_foam\_brick*          & 97.2  & 92.0  & 98.2  & \textbf{98.3}  \\  \hline
\hline
Average                    & 93.0  & 93.2   & 96.1  & \textbf{96.3}  \\  \hline

\hline

& \begin{tabular}{@{}c@{}}PoseCNN \\ (ICP)\end{tabular}  & \begin{tabular}{@{}c@{}}DF \\ (iterative)\end{tabular}  & \begin{tabular}{@{}c@{}}PVN3D \\ (ICP)\end{tabular} & \begin{tabular}{@{}c@{}}Ours \\ (depth)\end{tabular}  \\
\hline
\hline
\hline
002\_master\_chef\_can     & 68.1  & 73.2  & 79.3  & 70.9  \\  \hline
003\_cracker\_box          & 83.4  & 94.1  & 91.5  & 93.5  \\  \hline
004\_sugar\_box            & 97.1  & 96.5  & 96.9  & 97.9  \\  \hline
005\_tomato\_soup\_can     & 81.8  & 85.5  & 89.0  & 90.1  \\  \hline
006\_mustard\_bottle       & 98.0  & 94.7  & 97.9  & 91.1  \\  \hline
007\_tuna\_fish\_can       & 83.9  & 81.9  & 90.7  & 90.2  \\  \hline
008\_pudding\_box          & 96.6  & 93.3  & 97.1  & 97.9  \\  \hline
009\_gelatin\_box          & 98.1  & 96.7  & 98.3  & 98.3  \\  \hline
010\_potted\_meat\_can     & 83.5  & 83.6  & 87.9  & 85.1  \\  \hline
011\_banana                & 91.9  & 83.3  & 96.0  & 96.9  \\  \hline
019\_pitcher\_base         & 96.9  & 96.9  & 96.9  & 97.5  \\  \hline
021\_bleach\_cleanser      & 92.5  & 89.9  & 95.9  & 93.9  \\  \hline
024\_bowl*                 & 81.0  & 89.5  & 92.8  & \textbf{94.3}  \\  \hline
025\_mug                   & 81.1  & 88.9  & 96.0  & 95.3  \\  \hline
035\_power\_drill          & 97.7  & 92.7  & 95.7  & 93.7  \\  \hline
036\_wood\_block*          & 87.6  & 92.8  & 91.1  & \textbf{94.1}  \\  \hline
037\_scissors              & 78.4  & 77.9  & 87.2  & 90.2  \\  \hline
040\_large\_marker         & 85.3  & 93.0  & 91.6  & 93.6  \\  \hline
051\_large\_clamp*         & 75.2  & 72.5  & \textbf{95.6}  & \textbf{95.6}  \\  \hline
052\_extra\_large\_clamp*  & 64.4  & 69.9  & \textbf{90.5}  & 89.1  \\  \hline
061\_foam\_brick*          & 97.2  & 92.0  & 98.2  & \textbf{98.3}  \\  \hline
\hline
Average                    & 85.4  & 86.1  & 92.3  & \textbf{92.7}  \\  \hline

\hline
\caption{The quantitative evaluations on YCB-Video. Methods (DF(iterative) \cite{DenseFusionChen}, PoseCNN+ICP \cite{PoseCNN},  PVN3D+ICP~\cite{PVN3D2020Yisheng}) using depth information are compared. AUCs in ADD and ADD-S are computed. ``*": symmetric objects.}
\vspace{-2mm}
\label{Tab:YCB_res}
\end{longtable}
\endgroup

\subsection{Results on YCB-Video}
The YCB-Video dataset is introduced in~\cite{PoseCNN} which contains 21 objects with overall of 133,827 images. Following the dataset split in~\cite{PoseCNN}, the 80 videos with 80,000 synthetic images are for training and 2,949 keyframes are for testing. For convenience, both ADD and ADD-S AUC~\cite{PoseCNN} are used for testing.

As shown in Table~\ref{Tab:YCB_res}, AUC under ADD and ADD-S distance is evaluated for all the 21 objects in the dataset, YCB-Video. ADD-S is specifically designed for symmetric objects. Following settings in~\cite{PoseCNN, PVN3D2020Yisheng}, five symmetric objects are evaluated using ADD-S. For comparing with most of the best results on the YCB-Video dataset, we mainly refer to the RGBD methods or the RGB methods with ICP refinement which both consider the depth as the available information. The graph in our system is initialized by the points transformed from the depth map. We don't design any iterative refinement for our system. However, the proposed method reaches comparable performance with PVN3D+ICP~\cite{PVN3D2020Yisheng}. And it outperforms the DenseFusion~\cite{DenseFusionChen} method with iterative refinement. Since our method predicts the object surface at the same time while doing the keypoint estimation, more information is learned by our approach. We consider that the combination of the surface deformation and the keypoint prediction in multiple unpooling stages is like an increment refinement that is implicitly learned during the training stage. So we don't use an ICP for comparison here.

In the YCB-Video dataset, five objects are treated as symmetric objects, which are "024\_bowl", "036\_wood\_block", "051\_large\_clamp", "052\_extra\_large\_cl-amp" and "061\_foam\_brick". "025\_mug" is an object that is of the non-symmetric 3D model but may appear symmetry when some of its parts are occluded. With better rotation prediction, the proposed method obtains consistent performance improvement on these objects. 

\begin{figure*}[h]
    \centering
    \scalebox{0.7}{\includegraphics{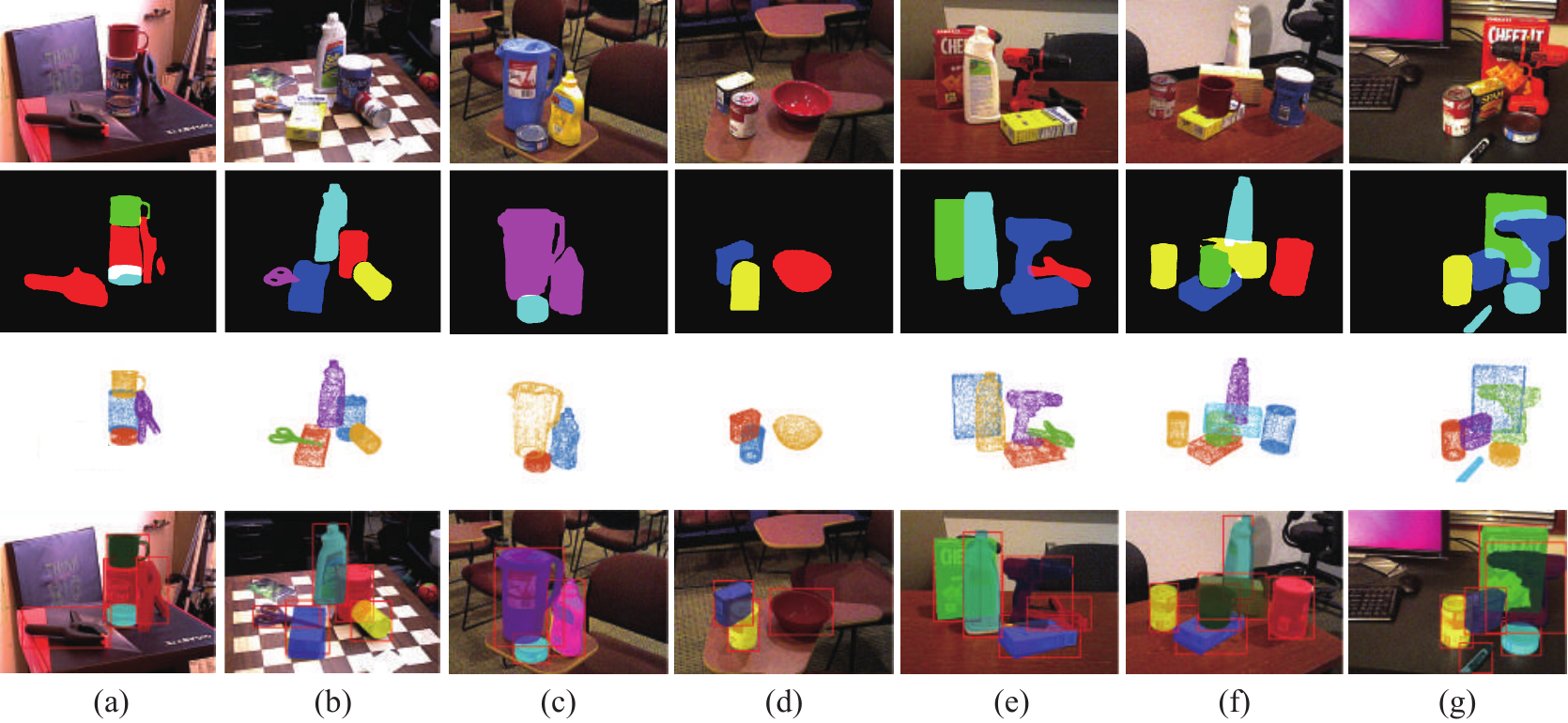}}
    \vspace{-2mm}
    \caption{Results visualization on the YCB-Video dataset. Images from top to bottom in each column are an original image, the instance segmentation result, the object pose visualization and the object pose visualization with ICP refinement, respectively, where the pose visualizations are generated by projecting 3D models to their 2D shapes.}
    \normalsize
    \label{fig:YCB_vis}
\end{figure*}

Another challenge of the YCB-Video dataset is the confusion of the object detector for the classification between "051\_large\_clamp" and "052\_extra\_large\_cl-amp" in the YCB-Video dataset. We give some visualization results in Fig.~\ref{fig:YCB_vis}. For Fig.~\ref{fig:YCB_vis} (a), two clamps are classified as ``051\_large\_clamp", in which one of them is wrongly classified. However, our result without ICP refinement shows proper pose prediction for the wrongly predicted class, which suggests that our method has learned the proper visual feature from the image appearance.

\subsection{Results on LINEMOD}
We evaluate the proposed system on the LIEMOD~\cite{ModelTextureStefan} dataset with and without the depth information. The LINEMOD dataset contains 13 objects in 13 videos in extreme background clutter. We create the training set following~\cite{PVNetSida} that crops object masks under ground-truth poses in each image and render them to different background images from PASCAL VOC~\cite{PascalVOCMark}. And the training and testing split are following~\cite{PoseCNN}.

\begingroup
\footnotesize
\centering 
\renewcommand\arraystretch{0.88}
\begin{longtable}{@{\extracolsep{\fill}}l|c|c|c|c|c}

\hline
& \multicolumn{5}{c}{RGB}  \\
\hline
& DeepIM  & PVNet  & CDPN  & Ours(plane) & - \\
\hline
\hline
ape            & 77.0  & 43.6  & 64.4  & 75.6   & -  \\  \hline
benchvise      & 97.5  & 99.9  & 97.8  & 93.2   & -  \\  \hline
camera         & 93.5  & 86.9  & 91.7  & 93.5   & -  \\  \hline
can            & 96.5  & 95.5  & 95.9  & 91.3   & -  \\  \hline
cat            & 82.1  & 79.3  & 83.8  & 81.2   & -  \\  \hline
driller        & 95.0  & 96.4  & 96.2  & 96.2   & -  \\  \hline
duck           & 77.7  & 52.6  & 66.8  & 73.1   & -  \\  \hline
eggbox         & 97.1  & 99.2  & 99.7  & 99.9   & -  \\  \hline
glue           & 99.4  & 95.7  & 99.6  & 100.0  & -  \\  \hline
holepuncher    & 52.8  & 82.0  & 85.8  & 83.3   & -  \\  \hline
iron           & 98.3  & 98.9  & 97.9  & 97.1   & -  \\  \hline
lamp           & 97.5  & 99.3  & 97.9  & 98.1   & -  \\  \hline
phone          & 87.7  & 92.4  & 90.8  & 91.1   & -  \\  \hline
\hline
Average        & 88.6  & 86.3  & 89.9  & \textbf{90.3} & -  \\  \hline

\hline

\hline
& \multicolumn{5}{c}{RGBD}  \\
\hline
& Point-Fusion  & \begin{tabular}{@{}c@{}}DF \\ (perpixel)\end{tabular}  & \begin{tabular}{@{}c@{}}DF \\ (iterative)\end{tabular}  & PVN3D  & \begin{tabular}{@{}c@{}}Ours \\ (depth)\end{tabular} \\
\hline
\hline
ape            & 70.4  & 79.5  & 92.3   & 97.3   & 96.3   \\  \hline
benchvise      & 80.7  & 84.2  & 93.2   & 99.7   & 99.7   \\  \hline
camera         & 60.8  & 76.5  & 94.4   & 99.6   & 99.6   \\  \hline
can            & 61.1  & 86.6  & 93.1   & 99.5   & 99.3   \\  \hline
cat            & 79.1  & 88.8  & 96.5   & 99.8   & 99.6   \\  \hline
driller        & 47.3  & 77.7  & 87.0   & 99.3   & 99.3   \\  \hline
duck           & 63.0  & 76.3  & 92.3   & 98.2   & 98.9   \\  \hline
eggbox         & 99.9  & 99.9  & 99.8   & 99.8   & 99.9   \\  \hline
glue           & 99.3  & 99.4  & 100.0  & 100.0  & 100.0  \\  \hline
holepuncher    & 71.8  & 79.0  & 92.1   & 99.9   & 99.9   \\  \hline
iron           & 83.2  & 92.1  & 97.0   & 99.7   & 99.7   \\  \hline
lamp           & 62.3  & 92.3  & 95.3   & 99.8   & 99.9   \\  \hline
phone          & 78.8  & 88.0  & 92.8   & 99.5   & 99.7   \\  \hline
\hline
Average        & 73.7  & 86.2  & 94.3   & \textbf{99.4}   & \textbf{99.4}   \\  \hline

\hline
\caption{The quantitative evaluations on the LINEMOD dataset. Methods only based on RGB image (DeepIM~\cite{DeepIMYiLi}, PVNet~\cite{PVNetSida}, CDPN~\cite{CDPN2019Zhigang}) and methods using the depth information (Point-Fusion~\cite{DenseFusionChen}, DF (perpixel)~\cite{DenseFusionChen}, DF (iterative)~\cite{DenseFusionChen}, PVN3D~\cite{PVN3D2020Yisheng}) are compared.}
\vspace{-2mm}
\label{Tab:LINEMOD_res}
\end{longtable}
\endgroup

As shown in Table~\ref{Tab:LINEMOD_res}, for the LINEMOD dataset, we evaluate the proposed method with and without the depth information. Initialized by a plane of average object depth in the dataset, the evaluation inference of the system does not need the depth map. Only the transformed object points are required for training supervision. Observing the performance decreasing when not using the depth map, we may consider that the system degenerates to the 2D coordinate map method. Without the depth information, the graph network predicts the translation parameter as well as the depth map. However, the graph is doing deformation in 3D space where the vertices relationship is captured. So, the proposed method still achieves comparable results with the state-of-the-art three methods~\cite{DeepIMYiLi, PVNetSida, CDPN2019Zhigang}. Since there is a gap between synthetic training samples, the system performance can be further improved by filling the gap.

In Table~\ref{Tab:LINEMOD_res}, we reach the state-of-the-art result comparing with other methods using the depth map. Most of our basic modules are from DF~\cite{DenseFusionChen}. After the graph matching and feature merging, the proposed method outperforms DF(per-pixel) and DF(iterative)~\cite{DenseFusionChen} over 10 and 4 percentage points respectively where we don't apply any iterative process. We also reach comparable performance with PVN3D~\cite{PVN3D2020Yisheng} which performs almost full correct rate under the ADD metric. However, we only do the segmentation from the 2D feature map while PVN3D~\cite{PVN3D2020Yisheng} does it in 3D space.

\subsection{Ablation Study}
In this section, we conduct experiments to test different modules and settings in the proposed system, \ie the spherical learning part and the graph regression. We compare different variants of our method by removing and modifying the modules in different ways.

~\RV{Before the ablation study on pose estimation, we first conduct an experiment on the rotated MNIST dataset to evaluate the effectiveness of spherical learning. Following~\cite{SphericalCNNTaco}, we rotate MNIST images on their spherical projection. Different combinations of non-rotated (NR) and rotated (R) sets for training and testing are considered. Similar to~\cite{SphericalCNNTaco}, we use a simple SphereCNN with two convolution layers. For learning the image rotation from the non-rotated training set, we do max-pooling on the features from the spherical correlation, which is different from~\cite{SphericalCNNTaco} that uses the $SO(3)$ integration to obtain the full rotation invariance. The pooled feature is considered to be rotation equivariance.}

\begin{table}[h]
\footnotesize
\begin{center}
\begin{tabular}{l|c|c|c}
\hline
training/testing  & NR/NR  & R/R  & NR/R  \\
\hline
\hline
flat convolution    &  0.98 &  0.17 &  0.10  \\ \hline
spherical learning  &  0.97 &  0.93 &  0.90  \\ \hline

\hline
\end{tabular}
\end{center}
\vspace{-5mm}
\caption{Classification accuracy results on the synthetic MNIST dataset. NR/R refers to that the models are trained on the non-rotated set and tested on the rotated set.}
\label{Tab:MNIST}
\end{table}

~\RV{Table~\ref{Tab:MNIST} reports similar results as those in~\cite{SphericalCNNTaco}. The pooled spherical features help obtain rotation-invariant results for different rotated inputs, which indicates the feasibility of our proposed approach to learn the rotation parameter from the spherical representation.}

\begin{table}[h]
\footnotesize
\begin{center}
\begin{tabular}{l|c|c|c|c|c|c}
\hline
    & MLP  & $G_{10}$  & $G_{20}$  & $G_{60}$  & $G_{120}$  & PVN3D\\
\hline
\hline
024\_bowl*                 & 90.5  & 85.4  & \textbf{94.3}  & 93.5  & 92.1  & 92.8  \\  \hline
036\_wood\_block*          & 89.0  & 83.5  & \textbf{94.1}  & 93.9  & 90.5  & 91.1  \\  \hline
051\_large\_clamp*         & 92.7  & 89.1  & 95.6  & \textbf{95.9}  & 94.6  & 95.6  \\  \hline
052\_extra\_large\_clamp*  & 86.5  & 90.3  & \textbf{91.1}  & 90.5  & 89.1  & 90.5  \\  \hline
061\_foam\_brick*          & 95.1  & 93.7  & \textbf{98.3}  & 97.7  & 97.5  & 98.2  \\  \hline

\hline
\end{tabular}
\end{center}
\vspace{-5mm}
\caption{Spherical learning ablation study.}
\label{Tab:Ablation_YCB}
\end{table}

First, we mainly test and compare the algorithm efficiency on symmetric objects in the YCB-Video that are "024\_bowl", "036\_wood\_block", "051\_large\_cl-amp", "052\_extra\_large\_clamp" and "061\_foam\_brick". Their AUCs under ADD-S are evaluated and the results are showing in Table~\ref{Tab:Ablation_YCB}.

In this experiment, we compare the spherical learning with direct regression by the Multi-Layer Perceptron (MLP). And the outputs with different grids are tested. The five symmetric objects in YCB-Video are used for testing. With all the other modules of the system remaining, only the spherical learning branch is changed. In Table~\ref{Tab:Ablation_YCB}, the ADD-S AUC results are shown. For all of the symmetric objects, using the spherical learning branch significantly improves the rotation regression.

We compare the results of our method under different output angle grid resolutions for the spherical learning branch. Particularly, we consider the grid sizes of 10, 20, 60 and 120 denote the corresponding variants of our method as $G_{10}$, $G_{20}$, $G_{60}$ and $G_{120}$, respectively. It can be seen that $20-60$ is a no-sensitive range of the output grid. Inside this grid range,  the performance of the spherical learning branch is not sensitive to its grid size. Thus, for all other experiments, we use $G_{20}$ for efficiency. Meanwhile, the results from PVN3D~\cite{PVN3D2020Yisheng} for the 5 symmetric objects are also compared which shows the efficiency of our proposed method on symmetric objects.

~\RV{Then, for better evaluating the performance of the proposed spherical learning module, we conduct the experiment combining the spherical learning standalone with the GCN branch. However, we use a keypoint-based regression instead of the GCN branch. The regression has the same output of the predefined keypoints.}

\begin{table}[h]
\renewcommand\arraystretch{0.88}
\footnotesize
\begin{center}
\begin{tabular}{l|c|c|c}
\hline
    & direct regression   & w/o sl*  & w/ sl* \\
\hline
\hline
YCB-Video           & 81.9  & 87.1  & 93.9  \\ \hline
LINEMOD             & 70.2  & 77.6  & 85.1  \\ \hline

\hline
\end{tabular}
\end{center}
\vspace{-5mm}
\caption{Ablation study for our method with/without the spherical learning combining the pose refinement modules. We report the ADD-S AUC on YCB-Video with depth input and LINEMOD without depth input over all the dataset objects. "sl*" represents the spherical learning module.}
\label{Tab:ablation_study}
\end{table}

~\RV{Table~\ref{Tab:ablation_study} shows the results of the ADD-S AUC on YCB-Video with depth input and LINEMOD without depth input over all the dataset objects. It can be seen that compared with the direct regression which directly regresses the pose parameter, both the spherical learning and the keypoint-based regression lead to better performance.}

\begin{table*}[h]
\renewcommand\arraystretch{0.88}
\footnotesize
\begin{center}
\begin{tabular}{l|c|c}
\hline
    & Coarse  & Refine  \\
\hline
\hline
YCB-Video (ADD-S)   & 91.1  & 96.3  \\ \hline
YCB-Video (ADD)   & 79.5  & 92.7  \\ \hline

\hline
\end{tabular}
\end{center}
\vspace{-5mm}
\caption{Ablation study for coarse rotation learning.}
\label{Tab:coarse}
\end{table*}

In Table~\ref{Tab:coarse}, we extract the rotation results from the coarse rotation estimation directly. From the results on YCB-Video, the performance deteriorates both under the ADD-S and ADD metrics. Due to the inaccurate of the coarse rotation estimation, the performance under ADD metric decreases a lot while the ADD-S result can maintain by satisfying translation estimation.

Then, we test different regression components of the proposed system. If both of the spherical learning and graph modules are removed, our method degrades to the baseline method, which directly regresses the pose parameters. For our method without the refinement module, the coarse rotation from the spherical learning is used as an incremental estimation that is added to the output of the baseline (direct parameter regression). For our method without the spherical learning part, we treat the output of the baseline as the coarse rotation estimation.

\begin{table*}[h]
\renewcommand\arraystretch{0.88}
\footnotesize
\begin{center}
\begin{tabular}{l|c|c|c|c}
\hline
    & \begin{tabular}{@{}c@{}}CNN \\ (parameter)\end{tabular}  & \begin{tabular}{@{}c@{}}CNN \\ (keypoints)\end{tabular}  & \begin{tabular}{@{}c@{}}graph \\ (parameter)\end{tabular}  & \begin{tabular}{@{}c@{}}graph \\ (keypoints)\end{tabular} \\
\hline
\hline
YCB-Video   & 85.9  & 93.6  & 88.3  & 96.3  \\ \hline
LINEMOD     & 71.1  & 85.9  & 83.2  & 90.3  \\ \hline

\hline
\end{tabular}
\end{center}
\vspace{-5mm}
\caption{Ablation study for the keypoint regression.}
\label{Tab:ablation_regression}
\end{table*}

In Table~\ref{Tab:ablation_regression}, the comparison using different regression components is illustrated. Following the dataset settings of previous experiments, we test the ADD-S AUC on YCB-Video with depth input and LINEMOD without depth input. Besides the keypoint prediction, direct pose parameter regression is compared. For the direct pose regression, we predict the 6D pose parameter on the output feature maps. After an average pooling, the final pose is obtained. From Table~\ref{Tab:ablation_regression}, the keypoint prediction improves the overall accuracy both on CNN and the graph module. Meanwhile, with the spatial projection, the graph module can further improve the performance.

\begin{table*}[h]
\footnotesize
\begin{center}
\begin{tabular}{l|c|c}
\hline
    & Closest matching  & bipartite graph matching  \\
\hline
\hline
YCB-Video   & 93.0  & 96.6  \\ \hline
LINEMOD     & 89.5  & 90.3  \\ \hline

\hline
\end{tabular}
\end{center}
\vspace{-5mm}
\caption{Ablation study for bipartite graph matching.}
\label{Tab:graph_matching}
\end{table*}

The last experiment in Table~\ref{Tab:graph_matching} demonstrates the function of the bipartite graph matching. Without the bipartite graph matching, the nearest assignment from the convolution feature to the graph introduces repeat and missing points which makes the feature assignment become aliasing in the spatial space. With a bipartite graph matching to make an average of the assignment provides better feature alignment which obtains better results as shown in Table~\ref{Tab:graph_matching}.

\section{Conclusion}
\label{sec:Conclusion}

Considering the strength of Graph Convolutional Network (GCN) in finding the relationships among vertices, we abstract the 3D mesh as the graph and use a deep neural network to predict keypoints during the inference. For better convergence of the GCN, the static graph is built and the Poisson disc sampling is used to extract 2D points. The Poisson disc sampling introduces randomness into the network training so that the final output is more robust. Furthermore, to reduce the inconsistency of the static graph and the random point sampling, a bipartite graph matching mechanism is introduced into our system. Meanwhile, the 3D shape is used as supervision in the proposed training scheme. The graph unpooling is utilized for better fitting the 3D shape label, which also helps the low-level feature forward and the feature voting.

We have explored the feasibility of using spherical learning to alleviate the rotation symmetry problem in the object 6D pose estimation task. A spherical representation of an object is a natural representation inheriting the symmetry property of the object. Our proposed approach estimates the object rotation based on the spherical correlation in deep neural networks with spherical learning. On two very challenging datasets, we have compared our method with the state-of-the-art methods. Our system achieves stable performance improvement for most of the objects, which demonstrates the feasibility and effectiveness of the proposed approach.


{\small
\bibliography{ref}
}

\end{document}